\pdfoutput=1

\documentclass[11pt]{article}

\usepackage[final]{acl}

\usepackage{times}
\usepackage{latexsym}

\usepackage[T1]{fontenc}

\usepackage[utf8]{inputenc}

\usepackage{microtype}

\usepackage{inconsolata}

\usepackage{graphicx}

\usepackage{amsmath}

\title{Valuable Hallucinations: Realizable Non-Realistic Propositions}

\author{Qiucheng Chen \\
  College of Intelligence and Computing \\
  Tianjin University \\
  \texttt{qiucheng\_chen@tju.edu.cn} \\\And
  Bo Wang \\
  College of Intelligence and Computing \\
  Tianjin University \\
  \texttt{bo\_wang@tju.edu.cn} \\}

\begin{document}
\maketitle
\begin{abstract}
This paper introduces the first formal definition of valuable hallucinations in large language models (LLMs), addressing a gap in the existing literature. We provide a systematic definition and analysis of hallucination value, proposing methods for enhancing the value of hallucinations. In contrast to previous works, which often treat hallucinations as a broad flaw, we focus on the potential value that certain types of hallucinations can offer in specific contexts. Hallucinations in LLMs generally refer to the generation of unfaithful, fabricated, inconsistent, or nonsensical content. Rather than viewing all hallucinations negatively, this paper gives formal representations and manual judgments of "valuable hallucinations" and explores how realizable non-realistic propositions—ideas that are not currently true but could be achievable under certain conditions—can have constructive value.

We present experiments using the Qwen2.5 model and HalluQA dataset, employing ReAct prompting (which involves reasoning, confidence assessment, and answer verification) to control and optimize hallucinations. Our findings show that ReAct prompting results in a 5.12\% reduction in overall hallucinations and an increase in the proportion of valuable hallucinations from 6.45\% to 7.92\%. These results demonstrate that systematically controlling hallucinations can improve their usefulness without compromising factual reliability.\footnote{The paper uses an AI assistant to refine the expression of certain sections, but the research and coding parts of the paper were entirely conducted without the use of AI.}
\end{abstract}

\section{Introduction}
\subsection{Background and Problem Statement}

In recent years, large language models (LLMs) \citep{12,13,14,15,16} have achieved remarkable progress in the field of natural language processing (NLP), significantly advancing capabilities in language understanding \citep{3,4}, generation \citep{5,6}, and reasoning \citep{7,8,9,10,11}. However, alongside these rapid advancements, a concerning issue has emerged: these models tend to generate hallucinations \citep{58,59,60}, content that appears plausible but is factually incorrect or unfaithful to the input \citep{61}. Hallucinations pose significant challenges in truth-sensitive domains such as finance \citep{51}, law \citep{52}, science \citep{53,54}, and education \citep{55}.

The prevailing view in existing research is that hallucinations are detrimental, as they undermine the reliability of LLMs \citep{2}. Consequently, numerous studies have focused on mitigating hallucinations through fact-centric metrics  \citep{44,45,46,47}, benchmarks \citep{48,49,50}, and retrieval-augmented generation (RAG) techniques\citep{47,65}. Despite these efforts, Banerjee et al. 
 \citep{27} and Xu et al. \citep{28} have demonstrated that hallucinations are inherent to LLMs, arising from their underlying mathematical and logical structures, and cannot be entirely eliminated through architectural improvements, dataset enhancements, or fact-checking mechanisms.

\subsection{Research Motivation and Limitations of Existing Work}

While most research treats hallucinations as entirely harmful, a small but growing body of work has begun to explore their potential value. For instance, Sui et al. \citep{43} suggest that hallucinations exhibit rich patterns of narrative behavior, while Wiggers \citep{75} refers to them as collaborative creative partners. In practical applications, Yuan et al. \citep{73} found that hallucinations can enhance the performance of LLMs in drug discovery tasks, and Wang \citep{74} demonstrated beneficial interactions between hallucinations and creativity in a multimodal AGI model. In scientific research, the creativity of LLMs has been shown to expand the boundaries of human knowledge and assist researchers in achieving breakthroughs \citep{76}.

However, existing studies on the positive effects of hallucinations are fragmented and lack a systematic definition or analysis. This paper aims to address this gap by introducing the concept of "valuable hallucinations" and providing a formal definition and classification framework.

\subsection{Core Contributions of This Work}

The core contributions of this paper are as follows:
\begin{itemize}
    \item Introducing the Concept of "Valuable Hallucinations": We formally define "valuable hallucinations" as realizable but non-realistic propositions. These propositions, if realized, could offer innovative and inspiring ideas, providing new perspectives or solutions to real-world problems.
    \item Systematic Classification and Analysis: Building on existing hallucination taxonomies (e.g., intrinsic-extrinsic dichotomy and factuality vs. faithfulness hallucinations), we identify which types of hallucinations can be valuable. We emphasize that realizable but non-realistic propositions fall under the category of "valuable hallucinations."
    \item Experimental Validation: We design a set of comparative experiments using the HalluQA dataset and the Qwen2.5 model. By employing prompt engineering and reflection techniques, we demonstrate that these methods can effectively control hallucinations and increase the proportion of valuable hallucinations in model outputs.
\end{itemize}

Future Research Directions: We propose potential methods for further controlling and utilizing hallucinations, such as combining retrieval-augmented generation (RAG) and meta-learning, providing a roadmap for future research in this area.

\section{Definitions}
\subsection{Hallucinations}
The term "hallucination" originates from the fields of pathology and psychology, where it refers to the perception of entities or events that do not exist in reality \citep{26}. In the context of natural language processing (NLP), hallucination in large language models (LLMs) typically refers to the generation of unfaithful, fabricated, inconsistent, or nonsensical content \citep{1}. Hallucinations occur when LLMs produce outputs that deviate from the input prompts or factual reality, often due to limitations in their training data or reasoning capabilities.

While hallucinations are generally considered harmful, this paper focuses on a specific subset of hallucinations that may have potential value, which we term "\textbf{valuable hallucinations}."

\subsection{Valuable Hallucinations}
The challenge of balancing creativity and factual accuracy in LLMs is a central issue in their development \citep{29,30}. While most research aims to mitigate or eliminate hallucinations, Banerjee et al. \citep{27} and Xu et al. \citep{28} have demonstrated that hallucinations are inherent to LLMs and cannot be entirely eradicated. Therefore, rather than attempting to eliminate hallucinations, we propose to identify and utilize their "valuable" aspects.

\subsubsection{Definition of Valuable Hallucinations}
We define valuable hallucinations as realizable but non-realistic propositions. These are propositions that, while not grounded in current reality, could be realized in the future and may offer innovative or inspiring ideas. The "value" of these hallucinations can be assessed through feedback, particularly human feedback, in reinforcement learning frameworks. The value of LLM outputs can be understood in two ways:

\textbf{Innovation and Inspiration}: Valuable hallucinations can propose innovative (and understandably unrealistic) propositions or inspire humans to formulate such propositions. For example, an LLM might generate a novel architectural design that does not currently exist but could be realized in the future.

\textbf{New Ideas and Solutions}: Valuable hallucinations can provide new ideas or solutions to realistic propositions. For instance, an LLM might suggest a creative approach to solving a scientific problem, even if the specific details are not yet feasible.

\subsubsection{Classification of Valuable Hallucinations}

To better understand valuable hallucinations, we classify them based on existing hallucination taxonomies:

\textbf{Intrinsic vs. Extrinsic Hallucinations} \citep{21,22,23,24}.
Intrinsic dichotomy is manifested when the output content of the LLM contradicts the input content (prompts), and when the output of the LLM cannot be verified from the source content, the situation is called extrinsic dichotomy. The "inability to verify" referred to here can also be called a \textbf{non-realistic proposition}, i.e., in most cases, it may be due to the fact that the LLM is making up completely fictitious numbers, references, or events. It is also possible that the big model generates what it "speculates" in the absence of obvious data and other support. Even though the model's "speculative" content may not be entirely correct or reasonable, it has a certain degree of \textbf{realizability}. For instance, if the LLMs output the architecture and drawings of a building that does not currently exist. If the content displayed by this architecture and drawings is realizable, then people can judge that this content has the characteristics of "realizable" and "non-realistic," and it can trigger the "realization" of them. It is a valuable hallucination to think about architecture and drawings. 

Under this classification, extrinsic hallucinations are more likely to be valuable, as they often involve creative or speculative content that could inspire new ideas.

\textbf{Factuality vs. Faithfulness Hallucinations} \citep{17}.
Factuality hallucination is divided into factual inconsistency and factual fabrication according to whether the generated factual content can be verified by reliable resources; faithfulness hallucination is divided into instruction inconsistency, context inconsistency and logical inconsistency according to the consistency of the generated content. Among them, factual fabrication refers to the situation where the output content of an LLM contains situations that cannot be verified on the basis of established knowledge of reality; under this categorization criterion,\textbf{ we consider factual fabrication to be the main way of generating valuable hallucinations}. For example, when we have a conversation with LLMs about a certain question, the content that the LLM answers is "fabricated" \citep{43}, i.e., this kind of content is non-realistic; and although it is not possible to verify that the LLM's answer to this question is correct, we can learn from the LLM's mindset and logic chain in answering the question, and then use it in other cases when we encounter the question. Although it is impossible to verify whether LLM's answer to this question is correct or not, we can learn from LLM's way of thinking and logical chain of answering this question, and then try to think and solve problems in a similar way when encountering other problems (i.e., with certain realizability). 

Among these, factual fabrication is the primary source of valuable hallucinations, as it involves generating novel content that, while not currently verifiable, may offer innovative insights.

\subsubsection{Examples of Valuable Hallucinations}
To illustrate the concept of valuable hallucinations, consider the following example:

\begin{itemize}
    \item \textbf{Question}: "Explain the use of CCCC formulas in chemistry."
    \item \textbf{LLM Response}: The model generates a detailed explanation of "CCCC formulas," which stand for "Concentration, Composition, Cross-Interaction, and Curvature." While these formulas do not exist in current scientific literature, the model's explanation is internally consistent and could inspire new research directions in thermodynamics or chemical engineering.
\end{itemize}

Due to space limitations, the complete question, LLM response, and our analysis can be found in Appendix \ref{sec:appendix}.

\subsection{Formal Definitions}
To formalize the concept of valuable hallucinations, we define the following sets:

$$ T = \{ \text{all propositions} \} $$

$$p = \{ \text{reality proposition} \} $$

$$q = \{ \text{realizable proposition} \} $$

$$\neg p \cap q = \{ \text{valuable hallucination} \}$$

Where:
\begin{itemize}
    \item $p \cup \neg p = T$ 
    \item $q \cup \neg q = T$ 
\end{itemize}

Here, $ \neg p $ represents non-realistic propositions, and $ q $ represents realizable propositions. The intersection of these two sets defines valuable hallucinations: propositions that are not currently realistic but could be realized in the future.

The "valuable" characteristic can be defined and judged by the feedback (especially human feedback) in reinforcement learning. The "value" of the output of a large-scale language model is twofold: on the one hand, it is to propose innovative (also understood as unrealistic) propositions or to give inspiration to human beings to propose such propositions; on the other hand, it is to provide possible new ideas or solutions to realistic propositions. 

\section{Methodology}
In this section, we outline the methodology used to explore and control hallucinations in large language models (LLMs), with a focus on increasing the proportion of valuable hallucinations. Our approach combines prompt engineering, reflection techniques, and other advanced methods such as retrieval-augmented generation (RAG) and meta-learning. The goal is not to eliminate hallucinations entirely but to control them in a way that maximizes their potential value. That is to say, we need to increase the proportion of "valuable hallucinations" in the hallucinations, not to increase the proportion of hallucinations in the LLM-generated content.

\subsection{Prompts and Reflections}

\subsubsection{Background Knowledge}

\textbf{Prompt engineering} is a core technique in Generative AI, aimed at improving the performance and output quality of LLMs by designing and optimizing natural language instructions or prompts. Effective prompt engineering requires a deep understanding of model behavior and the ability to guide LLMs to generate accurate and insightful outputs.

In the context of hallucinations, prompt engineering can be used to control and filter the content generated by LLMs. By designing prompts that encourage the model to display intermediate reasoning processes (e.g., Chain-of-Thought) and additional validation requirements \citep{67}, we can reduce the likelihood of the model generating unfaithful or fabricated content. For example, prompts that require the model to show its reasoning steps or cite relevant information can help the model self-check and reduce the probability of generating hallucinations.

\textbf{Reflection techniques} complement prompt engineering by allowing the model to learn from its mistakes and improve its performance over time. Reflection is particularly well-suited for tasks that require iterative experimentation \citep{20}, such as decision-making, reasoning, and programming. In the context of hallucinations, reflection can help the model filter and evaluate its own outputs, distinguishing between valuable and non-valuable hallucinations.

The reflection process involves several key steps:
\begin{itemize}
    \item Self-Assessment: The model assesses whether its generated content is a hallucination. If it is, the model further evaluates whether the hallucination is valuable.
    \item Feedback Integration: The model incorporates feedback, particularly human feedback, to adjust its parameters and preferences. For example, human raters can rank the model's outputs based on the following hierarchy: no hallucinations > valuable hallucinations > non-valuable hallucinations.
    \item Iterative Improvement: Through multiple rounds of learning and optimization, the model learns to prioritize valuable hallucinations and suppress non-valuable ones.
\end{itemize}
By combining prompt engineering and reflection techniques, we can create a more robust framework for controlling hallucinations. Prompt engineering guides the model's initial reasoning process, while reflection allows the model to iteratively improve its outputs based on feedback. This combination not only reduces the likelihood of generating non-valuable hallucinations but also increases the proportion of valuable hallucinations in the model's outputs.

\subsubsection{Experiments}

To test the effectiveness of prompt engineering and reflection techniques, we designed a controlled experiment using the HalluQA \citep{68} dataset and the Qwen2.5 model \citep{69}. The experiment consisted of two groups:

\begin{itemize}
    \item \textbf{Control Group}: The language model was prompted with a simple instruction: "Please answer the following question: ...". This group served as a baseline, where the model was not explicitly guided to display its reasoning process.
    \item \textbf{Experimental Group}: The language model was prompted using the ReAct  \citep{71} framework, which combines reasoning and acting. The prompt was designed as follows: "Please preface your answer by describing your thought process and indicating your confidence level in the answer, citing relevant information as a basis for your answer and ensuring that the answer is consistent with the actual facts. Please answer the following question: ...". This prompt encourages the model to show its reasoning steps, thereby reducing the likelihood of generating hallucinations.
\end{itemize}

The two groups used the same dataset, model, and other variables, with the only difference being the prompt design. The goal was to compare the proportion of valuable hallucinations and non-hallucinatory content between the two groups.

\begin{figure}[t]
  \includegraphics[width=\columnwidth]{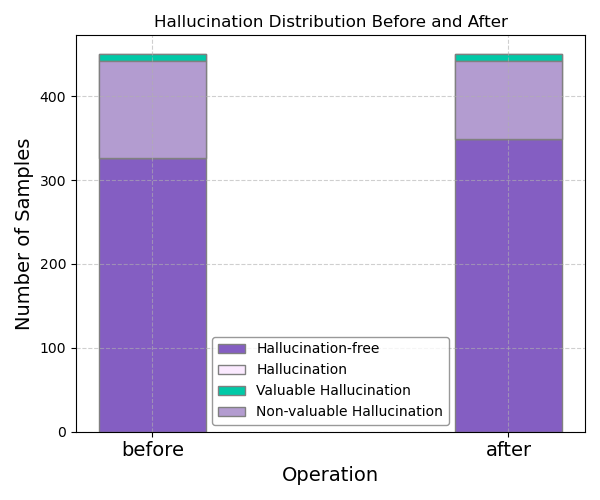}
  \caption{A figure with a comparison of the number of content types before and after ReAct prompts.}
  \label{fig:1}
\end{figure}

\begin{table*}
  \centering
  \begin{tabular}{lll}
        \hline
        \textbf{Type of Text} & \textbf{Normal prompts} & \textbf{ReAct prompts} \\
        \hline
        Non-hallucination & 326/450 & 349/450 \\
        \hline
        Hallucination & 124/450 & 101/450 \\
        \hline
        Subset of hallucination: valuable hallucinations & 8/124 & 8/101 \\
        \hline
        Subset of hallucination: non-valuable hallucinations & 116/124 & 93/101 \\
        \hline
  \end{tabular}
  \caption{\label{table:1}
    A table with a comparison of the number and percentage of content types before and after ReAct prompts.
  }
\end{table*}

The results showed that the experimental group (using ReAct prompts) had a higher proportion of valuable hallucinations and a lower proportion of non-valuable hallucinations compared to the control group. Specifically:

\begin{itemize}
    \item The proportion of valuable hallucinations increased by 1.47\% (6.45\%$ \rightarrow$7.92\%).
    \item The proportion of non-hallucinatory content increased by 5.12\% (72.44\%$ \rightarrow$77.56\%).
\end{itemize}

These results suggest that prompt engineering and reflection techniques can effectively control hallucinations and increase the proportion of valuable hallucinations in LLM-generated content.

Next, we use the Pearson correlation coefficient to calculate the correlation between the degree of hallucination of the output content after performing the prompting operation and the trust of the larger model in the answers it gives. Its formula is as follows: 

$$
r = \frac{\sum (X_i - \bar{X}) (Y_i - \bar{Y})}
{\sqrt{\sum (X_i - \bar{X})^2} \cdot \sqrt{\sum (Y_i - \bar{Y})^2}}
$$

For ease of calculation, we scored the content of the output of the large model to reflect its level of hallucination. The score for non-hallucinatory content was set to 2, valuable hallucinatory content was set to 1, and non-valuable hallucinatory content was set to 0. Also, those with a high level of trust were given a score of 2, those with a medium level of trust were given a score of 1, and those with a low level of trust were given a score of 0. The calculation tells us that r = 0.009, which is close to 0, indicating that there is almost no linear correlation between the degree of hallucination of the output content of the large model and its trust in the answers it gives.

Compare the outputs of the large model before and after the use of prompts, and observe the content of responses that were originally characterized as valueless hallucinations and were characterized as non-hallucinatory after the prompts were administered: 
\begin{itemize}
    \item The highest percentage of question areas (class) asked were pseudoscience at 17.5\%; myths and legends were next (12.5\%); and superstitions and geography types each accounted for 10\% of questions. 
    \item The Knowledge category of problems that tend to make factual errors in the large model is 35.0\%, the Misleading category of problems that tend to make Imitative Falsehoods \citep{49} in the large model is 47.5\%, and the Difficulty Misleading category of problems (Misleading-hard) is 17.5\%. 
\end{itemize}

\begin{figure*}[htbp]
    \centering
    \includegraphics[width=\textwidth]{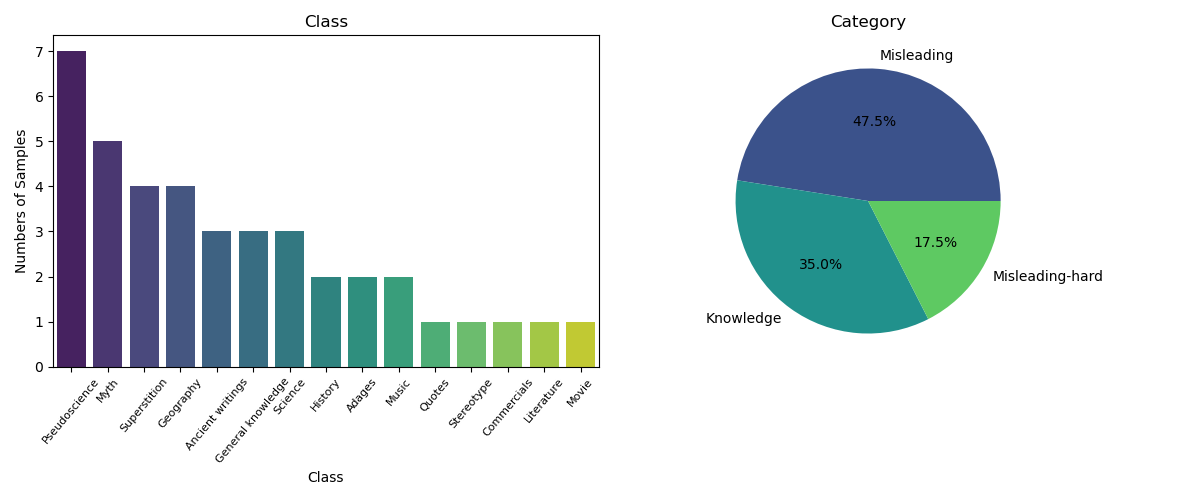} 
    \caption{The number and percentage of responses in the class and category to which the question belongs that originally manifested as a non-valuable hallucination and manifested as a non-hallucinatory response after prompting. }
    \label{fig:2}
\end{figure*}

Based on the above data, we find that prompting and reflection manipulations greatly reduce large model hallucinations caused by misleading type questions, while improving performance in pseudoscience, myths and legends domain quizzes. 

\subsection{Other Approaches to Control Hallucinations}

While prompt engineering and reflection techniques are effective in controlling hallucinations and increasing the proportion of valuable hallucinations, there are other advanced methods that could be explored to further enhance the control and utilization of hallucinations in large language models (LLMs). In this section, we discuss two promising approaches: retrieval-augmented generation (RAG) and meta-learning. Although we do not propose specific implementations in this paper, these methods offer potential directions for future research.

\subsubsection{Retrieval Augmented Generation (RAG)}
Retrieval-Augmented Generation (RAG) \citep{31,32,33} is a technique that integrates external information retrieval into the response generation process of LLMs. By searching external databases or knowledge graphs, RAG provides real-time contextual support to the generation process, significantly improving the factual accuracy and knowledge coverage of the model's responses.

In the context of hallucinations, RAG can be used to validate and refine the content generated by LLMs. For example, if the model generates a factual claim, RAG can retrieve relevant information from external sources to verify the claim's accuracy. If the claim is incorrect, the model can revise its response based on the retrieved information. This can help control hallucinations, increase the proportion of "valuable" hallucinations in hallucination content, and increase the rationality of LLM's innovative ideas.

Potential applications of RAG are as follows:
\begin{itemize}
    \item \textbf{Fact-Checking}: RAG can be used to fact-check the model's outputs in real-time, reducing the likelihood of generating non-valuable hallucinations.
    \item \textbf{Contextual Enrichment}: By retrieving relevant information from external sources, RAG can enrich the model's responses, making them more informative and accurate.
    \item \textbf{Iterative Refinement}: RAG can be integrated into a feedback loop, where the model iteratively refines its outputs based on retrieved information (e.g., the judgment of hallucination type), further improving the quality of its responses.
\end{itemize}

While we do not propose specific implementations in this paper, RAG offers a promising direction for future research in controlling hallucinations and increasing the proportion of valuable hallucinations.

\subsubsection{Meta-Learning}
Meta-learning, often understood as "learning to learn," refers to the process of improving a learning algorithm over multiple learning phases. In the context of LLMs, meta-learning can be used to fine-tune the model's parameters and output strategies to better adapt to specific tasks or domains.
Previously, many researchers have applied meta-learning techniques to NLP applications such as text categorization with excellent results. Meta-learning algorithms developed for image categorization can be applied to text categorization with only minor modifications to incorporate domain knowledge into each application \citep{37,38,39,40,41,42}. In the context of hallucinations, meta-learning could be used to categorize and filter the content generated by LLMs. For example, the model could be trained to recognize and prioritize valuable hallucinations while suppressing non-valuable ones.
Potential Applications of Meta-Learning are as follows:
\begin{itemize}
    \item \textbf{Fine-Tuning the Model}: By adjusting the model's parameters, learning rate, and number of training rounds, the model can be fine-tuned to specific domain types of text data, making it better adapted to the requirements of controlling hallucinations.
    \item \textbf{Adjusting Output Strategy}: Meta-learning could be used to adjust the model's output strategy, such as post-processing the model's output using regular expressions and other methods to reduce the hallucination of outputting valuable types.
    \item \textbf{Prompting and Guidance}: Meta-learning could be combined with prompt engineering to provide explicit instructions to the model, telling it to try to avoid outputting non-valuable hallucinations.
\end{itemize}

While we do not propose specific implementations in this paper, meta-learning offers a promising direction for future research in controlling hallucinations and increasing the proportion of valuable hallucinations. 

\section{Conclusion}
In this paper, we have explored the concept of \textbf{valuable hallucinations} in large language models (LLMs) and demonstrated that not all hallucinations are detrimental. By redefining hallucinations as \textbf{realizable but non-realistic propositions}, we have shown that certain types of hallucinations can provide innovative and inspiring ideas, offering new perspectives or solutions to real-world problems. Our work challenges the prevailing view that hallucinations are entirely harmful and provides a framework for identifying and utilizing their potential value.

\subsection{Key Contributions}
\begin{itemize}
    \item \textbf{Introduction of "Valuable Hallucinations"}: We formally defined valuable hallucinations as realizable but non-realistic propositions, providing a systematic framework for identifying and classifying them. This concept shifts the focus from eliminating hallucinations to leveraging their creative potential.
    \item \textbf{Experimental Validation}: Through a series of controlled experiments using the HalluQA dataset and the Qwen2.5 model, we demonstrated that prompt engineering and reflection techniques can effectively control hallucinations and increase the proportion of valuable hallucinations in LLM-generated content. Our results showed that these methods not only reduce non-valuable hallucinations but also enhance the model's ability to generate innovative and useful content.
    \item \textbf{Proposal of Advanced Methods}: We proposed additional methods, such as retrieval-augmented generation (RAG) and meta-learning, to further control hallucinations and increase the proportion of valuable hallucinations. These methods offer promising directions for future research and practical applications.
    \item \textbf{Practical Implications}: Our findings have significant implications for the use of LLMs in fields that require creativity and innovation, such as scientific research, artistic creation, and education. By controlling hallucinations and focusing on their valuable aspects, we can harness the full potential of LLMs while minimizing their risks.
\end{itemize}

\subsection{Final Thoughts}
In conclusion, this paper represents a significant step forward in understanding and utilizing hallucinations in LLMs. By redefining hallucinations as potentially valuable and providing methods to control and filter them, we have opened new avenues for research and application. Our work highlights the importance of balancing creativity and factual accuracy in LLMs and offers practical solutions for achieving this balance. We hope that this paper will inspire further research into the creative potential of LLMs and contribute to the development of more reliable and innovative AI systems.

\section{Limitations}

While this paper provides a foundation for understanding and utilizing valuable hallucinations in large language models (LLMs), there are several limitations that need to be acknowledged. These limitations highlight areas for future research and improvement.

\subsection{Dataset Scope and Model Constraints}
\begin{itemize}
    \item \textbf{Limited Dataset Scope}: HalluQA focuses primarily on structured question-answer pairs, which may not fully capture the diverse ways hallucinations manifest across different NLP tasks such as text summarization, open-ended reasoning, and dialogue systems.
    \item \textbf{Single Model Evaluation}: Our findings are specific to Qwen2.5, and the results may not generalize to other LLMs like GPT-4, LLaMA-2, or Claude. Future work should expand the scope to include multiple datasets (e.g., scientific literature, creative writing) and models (e.g., GPT-4, LLaMA) to validate the robustness and generalizability of our findings.
\end{itemize}

\subsection{Scope of Hallucination Classification}
Although we give a formal definition of valuable hallucinations, our classification remains somewhat subjective and context-dependent:
\begin{itemize}
    \item \textbf{Human Annotation Bias}: The determination of whether a hallucination is valuable involves subjective judgment \citep{57}, which could vary among different annotators.
    \item \textbf{Lack of Automated Metrics}: While we introduced trust consistency scores and human evaluation, there is no universally accepted automated metric to measure the usefulness of hallucinations. Future work could explore more robust computational frameworks for evaluation.
\end{itemize}

\subsection{Generalization Across Domains}
Our study primarily focuses on knowledge-based QA tasks, limiting its applicability to other domains:
\begin{itemize}
    \item \textbf{Scientific and Technical Domains}: The effectiveness of ReAct prompting in high-stakes fields such as healthcare, finance, or law remains uncertain. Misleading but plausible hallucinations could pose risks in these areas.
    \item \textbf{Creative Applications}: While valuable hallucinations are beneficial for fiction writing or brainstorming, their practical implications for scientific innovation and engineering design require further validation.
\end{itemize}

\subsection{Future Directions for Improvement}
In order to address these limitations, future research should:
\begin{itemize}
    \item Expand Model and Dataset Coverage: Test different LLMs and integrate broader datasets, including real-world, multi-domain corpora.
    \item Develop Automated Hallucination Metrics: Introduce scalable, objective scoring mechanisms for hallucination assessment.
    \item Optimize Prompting Efficiency: Explore alternative prompting methods, such as adaptive reasoning mechanisms that reduce response latency without sacrificing hallucination control.
\end{itemize}

By acknowledging these limitations, we provide a foundation for future work to enhance hallucination control and optimize the beneficial aspects of AI-generated content.

\bibliography{custom}

\appendix

\section{Appendix. Examples of Valuable Hallucinations}
\label{sec:appendix}
The following LLM Response is generated by ChatGLM \citep{77}.

\textbf{Question}: Please explain the use of CCCC formulas.

\textbf{LLM Response}: 

The "CCCC" acronym in the context of chemistry or chemical engineering typically stands for "Concentration, Composition, Cross-Interaction, and Curvature" formulas. These formulas are used to describe and predict the behavior of mixtures, particularly in the context of thermodynamics and phase equilibria. Here's a breakdown of each component:

Concentration: This refers to the amount of a substance in a given volume of the mixture, which can be expressed in various ways, such as molarity, molality, or mass fraction.

Composition: This is the relative amounts of each component in the mixture, usually expressed as mole fractions or weight fractions.

Cross-Interaction: This term refers to the interactions between different components in a mixture. In non-ideal mixtures, the behavior of one component can be affected by the presence of another component due to cross-interactions. These interactions are often quantified by parameters specific to the model being used.

Curvature: This term relates to the non-ideal behavior of mixtures, which can be quantified by parameters such as activity coefficients or the deviation from Raoult's law and Dalton's law. Curvature can also refer to the shape of phase diagrams, which can be curved due to non-ideal mixing.

CCCC formulas are used in the following ways:

Activity Coefficient Models: Models like the NRTL (Non-Random Two-Liquid), UNIQUAC (Universal Quasi-Chemical), or Wilson equation use CCCC-type formulas to predict activity coefficients, which account for the non-ideal behavior of liquid mixtures.
Equations of State: Advanced equations of state, such as the Peng-Robinson or Soave-Redlich-Kwong (SRK) models, incorporate CCCC principles to describe the thermodynamic properties of fluids, including mixtures, over a wide range of conditions.

Phase Equilibria Calculations: CCCC formulas are used to calculate phase equilibria, such as vapor-liquid equilibria (VLE), liquid-liquid equilibria (LLE), and solid-liquid equilibria (SLE), which are crucial for process design in chemical industries.

An example of a CCCC formula in action could be the following steps in the application of an activity coefficient model:

Concentration: Determine the concentration of each component in the mixture.

Composition: Calculate the mole fractions or weight fractions of the components.

Cross-Interaction: Use the model's parameters to describe the interactions between different components.

Curvature: Calculate the activity coefficients that account for the non-ideal behavior of the mixture.

For instance, in the NRTL model, the activity coefficient of a component i in a mixture is given by:

$$
\gamma_i = \exp\left[ \frac{\tau_{ij} G_j + \sum \left( x_k G_k \sum \left( \frac{\tau_{ik}}{\sum (\tau_{jk} x_j)} \right) \right)}{R T} \right]
$$

Where:
\begin{itemize}
    \item $\gamma_i$ is the activity coefficient of component $i$.
    \item $\tau_{ij}$ is the energy parameter that represents the cross-interaction between components $i$ and $j$.
    \item $G_j$ is the relative molal activity coefficient term.
    \item $x_k$ is the mole fraction of component $k$.
    \item $R$ is the gas constant.
    \item $T$ is the temperature in Kelvin.
\end{itemize}

The NRTL model is an example of a CCCC formula because it incorporates concentration (mole fractions), composition ($x_k$), cross-interaction ($\tau_{ij}$), and curvature (activity coefficients $\gamma_i$) to predict the behavior of liquid mixtures.

It's important to note that the specific equations and their applications can vary widely, and the interpretation of the CCCC acronym may differ depending on the context or the specific model being used. Always refer to the relevant scientific literature or textbooks for detailed information and equations.

\textbf{Our Analysis}:

In this experiment, we created a fabricated formula called "\textbf{CCCC}" and asked the large language model (LLM) to explain it. Both our question and the model's response are shown below. Upon review, it becomes clear that the model's answers reflect a type of \textbf{factual fabrication hallucination}.

Upon examining the LLM's explanation sentence by sentence, we find that each part of the explanation is internally consistent. There are no apparent theoretical errors in how the model explains the fabricated "CCCC" formula. The model correctly uses well-established chemical terms like \textbf{concentration}, \textbf{composition}, \textbf{cross-interaction}, and \textbf{curvature}, and appropriately references known variables in the formula. It even makes use of established models, such as the \textbf{NRTL model} (Non-Random Two-Liquid model) and equations that predict the behavior of liquid mixtures, all of which are commonly employed in thermodynamics and heat equilibrium analysis.

In other words, the LLM's response provides a technically coherent explanation for the invented "CCCC" formula. The generated answer, while based on our fabricated formula, draws upon existing theoretical knowledge and frameworks to describe how such a formula could potentially be used in practical scenarios.

One might argue that "the answer provided by the LLM does not introduce a new theorem; instead, it integrates existing variables and models used to describe liquid mixture behaviors, assigning them a new name based on the variables commonly considered in this type of analysis." This observation is correct, but we would like to address it from a different perspective.

While it is true that the model's response does not present groundbreaking or original theoretical advances (i.e., from "0 to 1"), the model's output is still "fabricated" and can stimulate innovative thinking. The answer is not based on any real or pre-existing "CCCC" formula but is instead a creative combination of familiar concepts. This fabricated content—though factually incorrect in the strict sense—has the potential to inspire new ideas or provide fresh ways of thinking about existing problems. Therefore, while it does not propose an entirely new theory, it does offer a valuable hallucination.

Thus, "valuable hallucinations" in LLMs are often the result of "fabrication"—the generation of new combinations or reinterpretations of existing knowledge, which may not be strictly factual but can spark new insights or lead to novel approaches. In this sense, hallucinations of this type are not only a form of fabrication but also hold potential for \textbf{generating creative and innovative ideas}, making them valuable in certain contexts.

\end{document}